# Pseudo vs. True Defect Classification in Printed Circuits Boards using Wavelet Features

Sahil Sikka · Karan Sikka · M. K. Bhuyan · Yuji Iwahori

**Abstract —** In recent years, Printed Circuit Boards (PCB) have become the backbone of a large number of consumer electronic devices leading to a surge in their production. This has made it imperative to employ automatic inspection systems to identify manufacturing defects in PCB before they are installed in the respective systems. An important task in this regard is the classification of defects as either true or pseudo defects, which decides if the PCB is to be re-manufactured or not. This work proposes a novel approach to detect most common defects in the PCBs. The problem has been approached by employing highly discriminative features based on multi-scale wavelet transform, which are further boosted by using a kernalized version of the support vector machines (SVM). A real world printed circuit board dataset has been used for quantitative analysis. Experimental results demonstrated the efficacy of the proposed method[1].

## Introduction

Printed circuit boards (PCB) are the basic building blocks in the consumer electronic industry. They are made from glass reinforced plastic with copper tracks instead of wires. Components are fixed in position by drilling holes through the board and then soldering them in place. The copper tracks link the components together forming a circuit. PCBs are rugged, inexpensive and highly reliable and so they are used in virtually all but the simplest commercially produced electronic devices. Automatic Optical Inspection (Iwahori et al. 2011, 2012) (AOI) system has been widely used to inspect defects in PCBs during the manufacturing process to avoid manufacturing of the defective PCBs in the consumer electronics market. Over the years PCBs have been forced to become dense and compact, and contain more circuits and components which are designed to implement more sophisticated tasks. This tendency in circuit layout makes the PCB inspection more problematic, leading to further challenges in developing advanced automated visual inspection systems for PCB.

Automatic optical inspection systems generally employ methods of Image Processing and Computer Vision, such as local feature matching, image skeletonisation, morphological image operations, which detect and classify the defects on the given PCBs. Although such methods have shown some success for "easy to spot" defects that often result from anomalies like bump or short in the circuit wiring, they are unable to truly discriminate between "hard to spot" defects like those resulting from production problems like oxidation, dust, contamination, etc. In this work the two classes, namely "easy to spot" and "hard to spot", have been defined as true and pseudo defects based on the nomenclature used in the previous studies by Iwahori et al. (2011,2012). One of the major drawbacks of previous works for this particular classification problem is high false positive rates, which reduces the practical utility of such systems. Reducing this false positives rate for true vs. pseudo defect classification is the prime motivation of this research work.

As mentioned above, the defects encountered in the PCBs are classified into two broad classes, namely true and pseudo defects. PCBs with true defects are not allowed to enter the market, whereas those with pseudo defects may be launched after some cleaning. True defects can be further classified into sub classes such as bump, broken piece, short, mousebite and particle (as shown in Fig. 1(a), Fig. 1(b), Fig. 1(c), Fig. 1(d) and Fig. 1(e) respectively). As mentioned earlier, most of these are physical defects and renders the PCB unusable. On the other hand, pseudo defects are temporary and occur in situations when dust particles stick to the electronic board or weak rust develops on the metallic parts of the PCB. Some examples of pseudo defects are shown in Fig. 2.

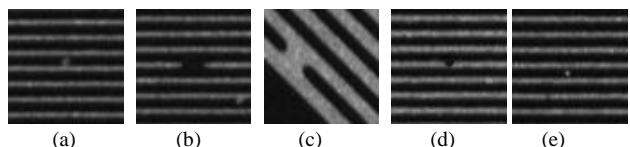

(a)  (b)  (c)  (d)  (e)
**Fig. 1 Examples of True defects (a) bump, (b) broken, (c) short, (d) mousebite, (e) particle**

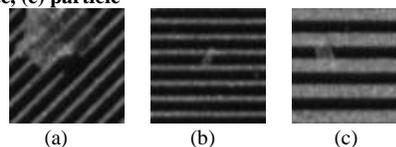

(a)  (b)  (c)
**Fig. 2 Examples of Pseudo defects**

## PREVIOUS WORK

Texture in image processing is defined as any region that is composed of small re-occurring pattern called texels. These texels can share some common properties like frequency information, directional derivatives, among others. Exploiting

Sahil Sikka is with Dhirubhai Ambani-Institute of Information and Communication Technology, Gandhinagar, Gujarat-382007, India (email: sahil_sikka@daiict.ac.in).

Karan Sikka is with University of California, San Diego, CA 92093, USA (email: ksikka@ucsd.edu).

M.K. Bhuyan is with Department of Electronics and Electrical Engineering, Indian Institute of Technology, Guwahati-781039, India (email: mkb@iitg.ernet.in).

Yuji Iwahori is with Department of Computer Science, Chubu University, Aichi 487-8501, Japan (email: iwahori@cs.chubu.ac.jp).

texture boils down to extracting meaningful properties from image regions that can encode their texture and later help in classification or segmentation. A number of techniques have been previously employed for texture, like higher order moments, Haralick features, Gabor wavelets, filter banks, texton etc (Haralick et al. 1973 and Manjunath and Ma 1996). Among these methods, features extracted using wavelet transform appears particularly interesting for this specific problem as it can extract information from data in multiple resolutions or scales. It has been shown in the experiments that simple statistical measures like first or second order moments over multi-resolution wavelet features provide considerably higher discriminative power. Such discriminative power is not only essential for the task of true vs. pseudo defect classification, but also entails a lower false positive rate than earlier approaches.

A significant work has been reported which is entirely based on using Hausdorff distance for image alignment and defect detection by Chen et al. 2005. For image alignment, a course-to-fine search technique is applied to minimize the time taken to calculate the Hausdorff distance between the reference and the inspected image. For defect detection, Hausdorff distance is calculated for every pixel in the inspected image and then compared with a predefined threshold. If the calculated distance is greater than the threshold, that pixel is labeled as a defect. For defect classification, local image features are extracted and passed to the support vector machine (SVM) for training and identifying defect types (Chen et al. 2005).

Another approach detects the defect region using image subtraction techniques by Iwahori et al. 2011. The features extracted from the resultant image were mean intensity, maximum intensity, minimum intensity, percentage of high intensity pixels, variance of the intensity and geometrical features like degree of circle and aspect ratio. These features were considered based on the fact that sizes of the defects are different for true and pseudo defects. The best results were obtained by modeling the SVM with mean intensity, minimum intensity and percentage of high intensity features. The method experimentally determines the best value of the parameters of the hard margin SVM with the evaluation equation which minimized the misjudged number of samples of true defect to pseudo defect. But, the problem is that the data plotted inside the margin are still unclear to be satisfied.

The above algorithm was improved by extracting proper shape and texture features Iwahori et al. 2012. In this, 9 shape features such as complexity, area of the defect, circularity, aspect ratio, minor axis length, major axis length, perimeter of the defect, diameter of the defect and the Euler number were extracted from the result of image subtraction. Also, 7 texture features like average grey level, entropy, standard deviation, smoothness, third moment, uniformity and grey level ratio were extracted from the defected image. The SVM gave best results when it was modeled with gray level ratio, smoothness of the defect, standard deviation and entropy (Iwahori et al. 2012). Though the automatic PCB defect classification is the most important and open research issues for VLSI industries, not much vision-based approaches are reported till date of this research work.

To address some of these issues, a novel scheme for PCB defect classification is proposed. The proposed algorithm uses a two-dimensional wavelet transform to segment the textural content of the image. Subsequently, statistical features like mean and standard deviation are calculated for two-level decomposition. This is then followed by modeling the SVM classifier for classifying the defects as true or pseudo.

**Proposed Method**

As explained earlier, the proposed method mainly depends on the analysis of texture pattern of the PCBs. For this, we propose to use Discrete Wavelet Transform (DWT), which is applied on each images of the dataset. DWT is a versatile approach for multi-resolution representation of signals. Compared to Fourier transform it has the advantage of being localized in both time and frequency domain. DWT is calculated using a filter bank consisting of a set of quadrature mirror filters comprising of both low and high-pass filters. Application of DWT to an image produces four components. The output of the low pass filter is called approximated image and those of the high pass filter are detailed regions of the image which can be interpreted as edges along vertical, horizontal and diagonal directions (Singh et al. 2011). The 2-D DWT decomposition is illustrated in Fig. 3. For each level, the input signal is filtered along the rows and the resultant signal is filtered along the columns (Zerves et al. 2001, Arivazhagan and Ganeshan 2003).

As shown in Fig. 3, with $L$ representing the level of decomposition, the 2D decomposition of an input image $IN[M][N]$ with $M$ columns and $N$ rows, $N_H$ and $N_L$ representing the number of taps of high pass and low pass filters whose impulse responses are $h_n$ and $w_n$ respectively is described by the following equations:

$$L_{j+1}[row][m] = \sum_{i=0}^{N_L-1} w[i] \times LL_j[row][2m-i] \quad (1)$$

$$H_{j+1}[row][m] = \sum_{i=0}^{N_H-1} h[i] \times LL_j[row][2m-1-i] \quad (2)$$

$$LL_{j+1}[n][col] = \sum_{i=0}^{N_L-1} w[i] \times L_j[2n-i][col] \quad (3)$$

$$LH_{j+1}[n][col] = \sum_{i=0}^{N_H-1} h[i] \times L_j[2n-1-i][col] \quad (4)$$

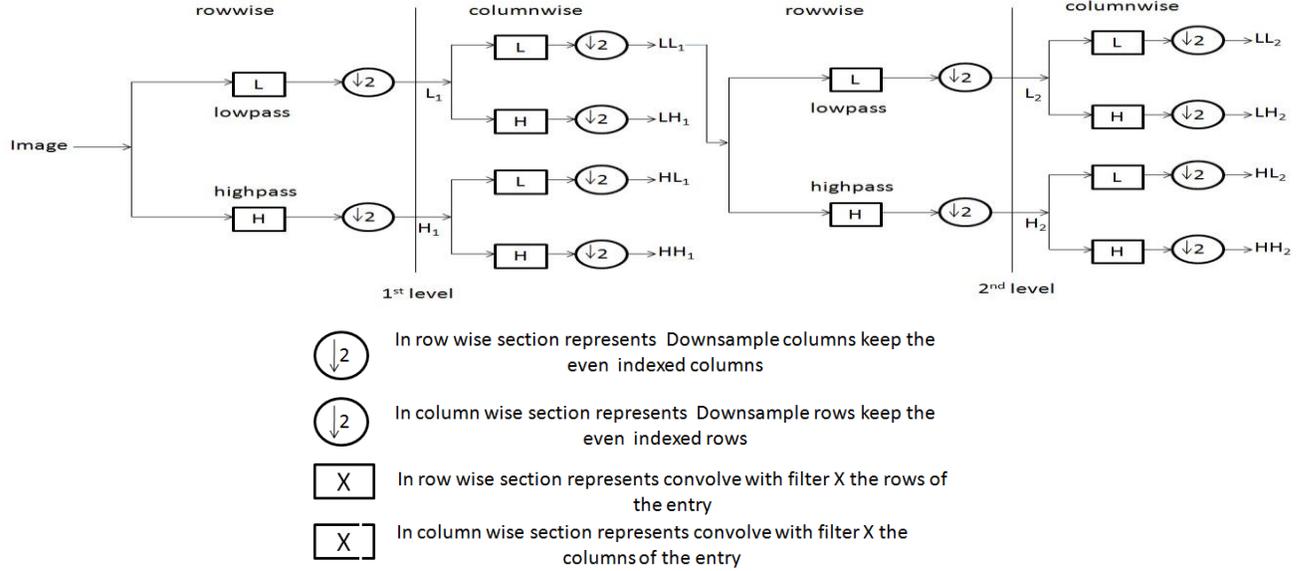

Fig. 3 Binary tree representation of 2-D DWT

$$HL_{j+1}[n][col] = \sum_{i=0}^{N_L-1} w[i] \times H_j[2n-i][col] \quad (5)$$

$$HH_{j+1}[n][col] = \sum_{i=0}^{N_H-1} h[i] \times H_j[2n-1-i][col] \quad (6)$$

In this, the terms $LL_{j+1}[n][col]$, $LH_{j+1}[n][col]$, $HL_{j+1}[n][col]$, $HH_{j+1}[n][col]$ are the approximated image, detailed images in the vertical, horizontal and the diagonal directions respectively and $j = \{0,1,\ldots L-1\}$, $row = \{0,1,\ldots N/2^j - 1\}$, $m = \{0,1,\ldots,2^{j+1}-1\}$, $col = \{0,1,\ldots M/2^{j+1}-1\}$, $n = \{0,1,\ldots N/2^{j+1}-1\}$ and $LL_0[n][m] = IN[n][m]$.

The four sub-bands obtained by applying DWT decomposition on an image are shown in Fig. 4(a). The sub-bands labeled $LH_1$, $HL_1$ and $HH_1$ represent the finest scale wavelet coefficients i.e., detail images, whereas the sub-band $LL_1$ corresponds to coarse level coefficients i.e., approximated image. To obtain the two-level and three-level wavelet decompositions, the approximated image i.e., $LL_1$ and $LL_2$ are decomposed and critically sampled respectively. The two-level wavelet decomposition is shown in Fig. 4(b). The values or transformed coefficients in approximation and detail images (sub-band images) are important since they characterize the texture in the image. Statistical features such as mean and standard deviation are extracted from the approximation and detail sub-bands of one, two and three level decomposed images (i.e., $LL_k$, $LH_k$, $HL_k$ and $HH_k$ for $k=1,2,3$) using the formulas given in (7) and (8) respectively. This is done to identify the combination of levels that gave maximum accuracy. It is to be noted that the classical texture representation technique like co-occurrence features viz., entropy and contrast cannot give good results due to the fact that images under consideration are quite small, and thus have less discriminative information.

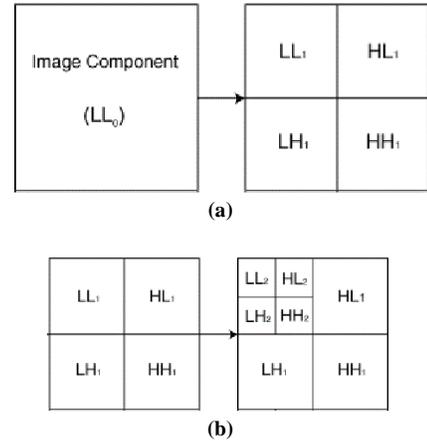

Fig. 4 Image decomposition: (a) one-level, (b) two-levels

$$mean(m) = \frac{1}{N^2} \sum_{i,j=1}^{N} p(i,j) \quad (7)$$

$$Standard\ deviation\ (S.D.) = \sqrt{\frac{1}{N^2} \sum_{i,j=1}^{N} \left[p(i,j)-m\right]^2} \quad (8)$$

where $p(i, j)$ is the transformed value in $(i, j)$ for any sub-band of size $N \times N$.

The statistical features extracted from the output images of the wavelet decomposition were then passed to the support vector machine (SVM) for training. SVM is a powerful binary classifier that maximizes the margin between the two classes (Iwahori et al. 2011, 2012). Given a set of training examples, each marked as belonging to one of the two classes, the SVM training algorithm builds a model that is employed to map novel examples into one of the two categories. For the present case, the two classes are true and pseudo defect.

For SVM, the discrimination function is written as

$$y(x) = w^t x + b \quad (9)$$

Using dual of (9), classification rule can be expressed as a linear combination of training examples (in terms of kernel function $K$) that is useful for applying the kernel trick using the kernel function in (11).

$$y(x) = \sum_{n \in S} a_n\, t_n\, K(x, x_n) + b \quad (10)$$

where, $w$ represents weight vector, $x$ is input data, $x_n$ represents the support vector, $a_n$ represents Lagrange multipliers, $t_n$ represents the teaching signal, $K(x,x_n)$ represents the kernel function, $b$ represents the threshold value and $S$ represents the set of support vectors.

Kernel trick makes it possible to perform dot products with the training data in higher dimensions that is able to separate data which is non-linearly separable in lower dimensions (non-linear classification). It allows calculating the dot products in higher dimensions without explicitly projecting the points. The proposed method employs the Gaussian Kernel, which is defined as:

$$K(x, x_n) = \exp\left(-\frac{\|x - x_n\|^2}{\sigma^2}\right) \quad (11)$$

**Experimental results**

**Dataset:** To validate the proposed algorithm, experiments were performed on the real electronic image dataset having *51* true defect images and *50* pseudo defect images each of size $64 \times 64$ pixels. Features were extracted from the images in this dataset using the algorithm discussed in section III. Fig. 5 shows a sample PCB and Fig. 6 shows some images of the database.

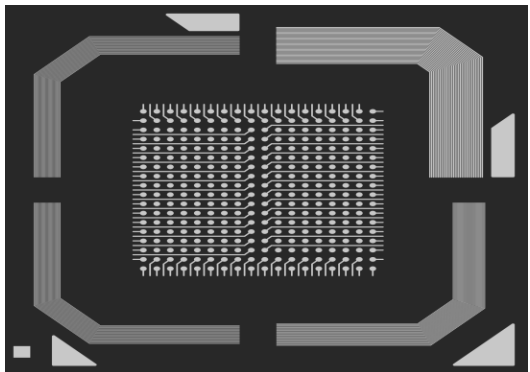

**Fig. 5  A sample PCB used in our experiments**

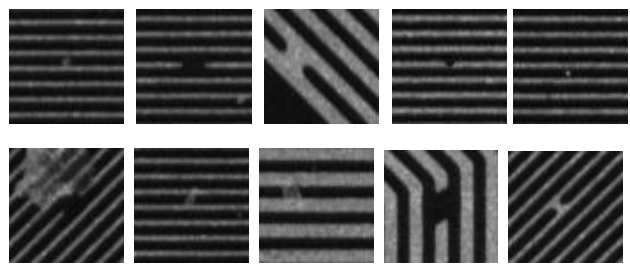

**Fig. 6  A part of the image database**

**Classifier:** The proposed classifier is trained using *26* true defect images and *24* pseudo defect images. The testing set consisted of *25* true images and *26* pseudo defect images. RBF kernelized SVM is used and value of the parameters ($\sigma$ and cost parameter '$c$') are empirically selected for maximum accuracy. Fig. 7 shows the 2-D discrete wavelet transforms of a pseudo defect image and Fig. 8 shows a true defect image.

The steps involved in feature extraction from the images are shown in Fig. 9.

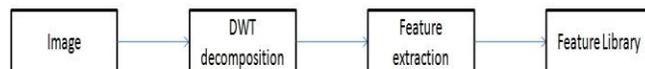

**Fig. 9  Feature extraction method**

TABLE I depicts correct classification rate i.e. *86.27%* using wavelet statistical features of *1*-level, *2*-level and *3*-level wavelet decompositions. It is evident that *2*-level decomposition proved to be the best among different levels tested giving maximum accuracy i.e. *86.27%* for $\sigma=0.01$ and cost factor '$c$' = *9*.

TABLE II, TABLE III and TABLE IV respectively show the confusion matrix of 1-level, 2-level and 3-level wavelet decomposition for the predicted values where each column represents the number of instances in their respective predicted classes, whereas each row represents number of instances in their respective actual classes. The inference that can be drawn looking at these tables is that 1-level and 3-level decompositions deliver better accuracy results for true defect classifications, but poor for pseudo defect classifications, whereas 2-level DWT yields good results for pseudo defect classification and comparable results for true defect classifications.

**Discussion**

The textures are true characteristics of a region, because they give us information about the spatial arrangement of color or intensities in an image. Thus methods like multi-resolution wavelet texture features are indeed good for classification of PCB defects into true and pseudo defects. The experiment was also performed using gray level ratio and Gabor filter, but promising results were achieved only using wavelet features. Gabors are not rotation invariant and the outputs of Gabor filter bank are not mutually orthogonal, which may result in significant correlation between texture features. Wavelet transform provides a precise and unifying framework for the analysis and characterization of a signal at different scales. Advantage of wavelet transform over Gabor filter is that the low pass and high pass filters used in the wavelet transform remain the same between two consecutive scales, whereas the Gabor approach requires filters of different parameters. In other words, Gabor filters require proper tuning of filter parameters at different scales. It is observed that false positive rate is high with *1*-level and *3*-level DWT. With *2*-level DWT, true negative rates are high and also, comparable true positive rates are obtained.

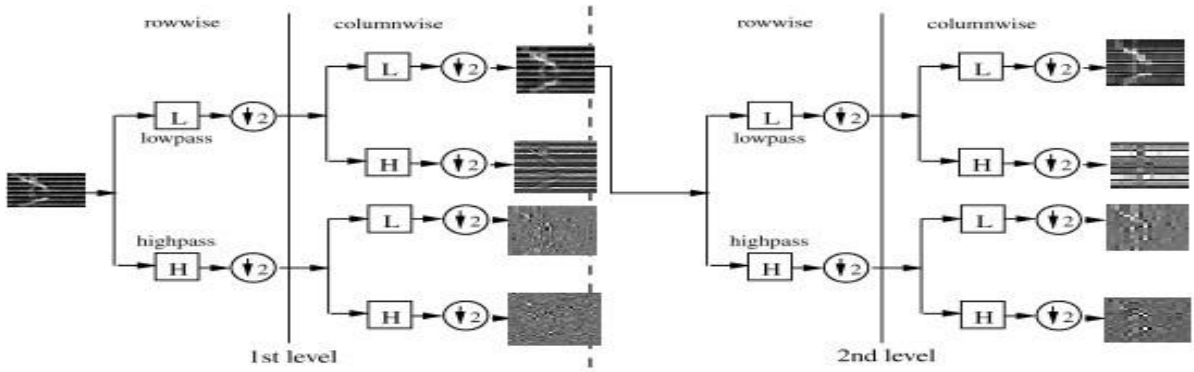

**Fig. 7 2-D DWT of a pseudo defect image**

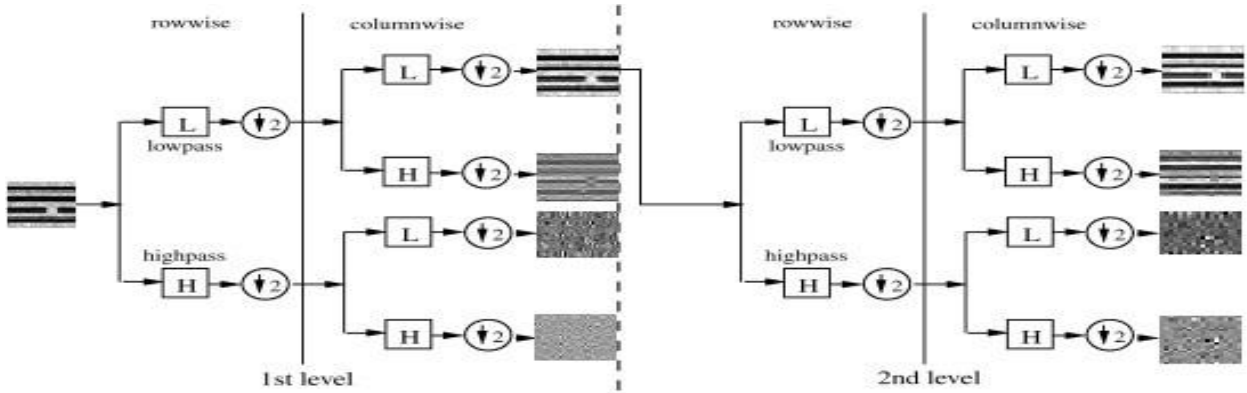

**Fig. 8 2-D DWT of a true defect image**

TABLE I
PERFORMANCE OF CLASSIFIER FOR DIFFERENT LEVELS OF WAVELET DECOMPOSITION

| $\sigma$ | c | 1-level Decomposition | 2-level Decomposition | 1-level Decomposition |
|---|---|---|---|---|
| 0.01 | 3 | 70.59% | 80.39% | 62.74% |
| 0.01 | 5 | 72.55% | 84.31% | 62.74% |
| 0.01 | 9 | 70.59% | 86.27% | 62.74% |
| 0.01 | 11 | 70.59% | 82.35% | 62.74% |
| 0.01 | 15 | 70.59% | 82.35% | 62.74% |
| 0.02 | 9 | 74.51% | 74.50% | 58.82% |
| 0.06 | 9 | 64.71% | 68.62% | 58.82% |
| 0.06 | 1 | 70.59% | 60.78% | 56.86% |
| 0.15 | 3 | 70.59% | 62.74% | 56.86% |

TABLE II
CONFUSION MATRIX FOR 1-LEVEL DECOMPOSITION

|  | True defects | Pseudo defects | Correct Classifications |
|---|---|---|---|
| True defects | 21 | 4 | 84% |
| Pseudo defects | 9 | 17 | 65.38% |

TABLE III
CONFUSION MATRIX FOR 2-LEVEL DECOMPOSITION

|  | True defects | Pseudo defects | Correct Classifications |
|---|---|---|---|
| True defects | 19 | 6 | 76% |
| Pseudo defects | 1 | 25 | 96% |

TABLE IV
CONFUSION MATRIX FOR 3-LEVEL DECOMPOSITION

|  | True defects | Pseudo defects | Correct Classifications |
|---|---|---|---|
| True defects | 25 | 0 | 100% |
| Pseudo defects | 19 | 7 | 26.92% |

## Conclusion

A novel methodology is devised for a specific automatic inspection system that classifies the manufacturing defects in a printed circuit board as either true or pseudo defect. The algorithm shows promising results when applied on a real PCB image dataset and hence can prove to be of significant utility in segregating PCBs with true or pseudo defect in real-life setting. The proposed scheme can be directly applied in the VLSI industries.

The algorithm begins with the segmentation of the image using 2-D wavelet transform upto *2*-levels. It is then followed by extracting statistical features like mean and standard deviation from the decomposed images. These features are then passed to RBF kernelized SVM classifier to classify the dataset into the appropriate classes i.e. true or pseudo.

The proposed system greatly reduces the false negative and false positive rates. In conclusion, this paper supports the claim that *2*-level wavelet statistical features can provide performance benefit for defect classification task on PCBs. The proposed algorithm is concocted keeping in mind of a fully automated system, thereby limiting the possible human intervention to a minimum level. The defective PCBs which form an essential component of the consumer electronic industry can thus be checked from being manufactured into the consumer electronics market.

## BIOGRAPHIES

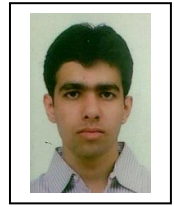

**Sahil Sikka** is currently in $3^{rd}$ year pursuing his Bachelor's in Information and Communication Technology (ICT) DA-IICT, Gujarat, India. His research interests include digital Image processing, Machine Learning and pattern recognition.

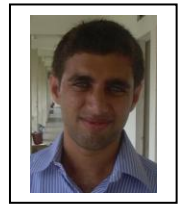

**Karan Sikka** completed his Bachelor's in Electronics and Communication Technology from Indian Institiute of Technology Guwahati, India in 2010. He is currently a Phd candidate at University of California, San Diego. His research intersets include computer vision and machine learning.

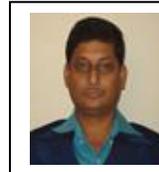

**M.K. Bhuyan** (M'10) is currently working as an Assistant Professor in the Department of Electronics and Electrical Engineering, IIT Guwahati, India. He has a Ph.D. degree in Electronics and Communication Engineering from Indian Institute of Technology Guwahati, India. He did his post-doctoral research in the school of Information Technology and Electrical Engineering, University of Queensland, Australia and then, in National ICT, Brisbane, Australia. He was also working as an Assistant Professor in the department of Electrical Engineering, IIT Roorkee, India. His broad research interests include Image/Video Processing, Computer Vision and Human Computer Interactions (HCI).

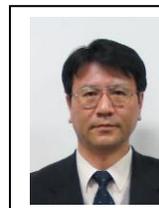

**Yuji Iwahori** received Ph.D. from the Department of Electrical and Electronics, Tokyo Institute of Technology in 1988. He joined Educational Center for Information Processing, Nagoya Institute of Technology as a research associate in 1988 and he became a professor of Center for Information and Media Studies, Nagoya Institute of Technology in 2002. Since 2004, he has joined Chubu University as a professor and he is currently the head of Graduate Program of Computer Science, Chubu University, Japan. In the meanwhile, he has been a visiting researcher of UBC Computer Science, Canada. His research interests include Computer Vision, Image Recognition and Artificial Intelligence. He got KES2008 Best Paper Award and he is a member of IEEE, IEICE, IPSJ and KES International.